# Hierarchical Abstraction Enables Human-Like 3D Object Recognition in Deep Learning Models

**Shuhao Fu (fushuhao@g.ucla.edu)**
Department of Psychology, UCLA

**Philip J. Kellman (kellman@psych.ucla.edu)**
Department of Psychology, UCLA

**Hongjing Lu (hongjing@g.ucla.edu)**
Department of Psychology, UCLA

## Abstract

Both humans and deep learning models can recognize objects from 3D shapes depicted with sparse visual information, such as a set of points randomly sampled from the surfaces of 3D objects (termed a *point cloud*). Although deep learning models achieve human-like performance in recognizing objects from 3D shapes, it remains unclear whether these models develop 3D shape representations similar to those used by human vision for object recognition. We hypothesize that training with 3D shapes enables models to form representations of local geometric structures in 3D shapes. However, their representations of global 3D object shapes may be limited. We conducted two human experiments systematically manipulating point density and object orientation (Experiment 1), and local geometric structure (Experiment 2). Humans consistently performed well across all experimental conditions. We compared two types of deep learning models, one based on a convolutional neural network (DGCNN) and the other on visual transformers (point transformer), with human performance. We found that the point transformer model provided a better account of human performance than the convolution-based model. The advantage mainly results from the mechanism in the point transformer model that supports hierarchical abstraction of 3D shapes.

## Introduction

Objects in the natural world are three-dimensional and possess physical properties such as geometric shape, material, and volume. From a brief glance, the human visual system excels in extracting visual attributes from pixel-level information in images to infer these object properties. Among the object properties, recognizing the three-dimensional shape of objects is considered fundamental for most daily life tasks involving navigation and interaction with the external world. Humans utilize depth cues and spatial relationships to understand and identify objects. Considerable research (Wallach & O'Connell, 1953; Marr, 1982; Liu, 1998) indicate that humans construct mental representations of objects that incorporate 3D structural information, facilitating recognition from different perspectives. Human 3D object recognition is both accurate and remarkably robust. A key example of this robustness is the ability to perceive and interpret 3D objects even with limited information, such as point cloud displays that are sparsely sampled along object surfaces (Treue, Husain, & Andersen, 1991; Treue, et al., 1995; Murray, Olshausen, & Woods, 2003; Wagemans, et al., 2012; Guo et al., 2020). Such high accuracy and robustness of human vision system appears particularly tuned to 3D object recognition. Previous research shows that 3D shape perception arises early in human infancy (Kellman, 1984), and that human toddlers begin to develop a strong shape bias between 18 and 24 months, shifting from an early reliance on texture and other perceptual features to generalizing object names based on shape as their vocabulary expands (Yee, Jones, & Smith, 2012). While sensitivity to texture evolves, it is secondary compared to shape-based object recognition.

These empirical evidence in human development is in contrast with the default strategies used in deep convolutional neural networks (DCNN) (Krizhevsky et al., 2012; Simonyan and Zisserman, 2014; He et al., 2015). Baker et al. (2018) found that DCNNs struggled to classify objects in images based on their global shape. In one experiment, they presented CNNs with object that preserved the global shape but were filled with textures from other objects. The networks showed a strong bias for classifying based on textures rather than shapes. Further experiments revealed that CNNs could not reliably classify objects based on outlines alone, indicating a reliance on local features rather than global shapes. These findings suggest that while CNNs can access local shape features in images, they do not form global shape representations crucial for human-like object recognition.

While the challenges of object recognition in 2D images are well-documented, the transition to 3D object recognition in the point cloud display introduces additional complexities and opportunities for both human and machine perception research. Recent advancements in deep learning have led to novel architectures designed specifically for 3D point clouds. PointNet, for example, introduced innovative ways to directly process point clouds by learning spatial features from raw 3D data (Qi et al., 2017a). Another significant development is Dynamic Graph CNN (DGCNN), which builds on the idea of graph neural networks to construct local neighborhoods and learn local geometric features dynamically (Wang et al., 2019). These models have shown human-like performance in various 3D recognition tasks, but whether they acquire 3D representations similar to humans remains unknown. The most recent advancement is to adopt the visual transformer architecture in 3D object recognition, such as the Point Transformer (Zhao et al., 2021). These transformer-based models also reach human-level recognition performance for 3d object recognition.

In this paper, we aim to compare the 3D object recognition in humans and deep learning models. Through a set of experimental manipulations, we analyze how both humans and models recognize 3D objects, particularly in challenging conditions where local features are disrupted or the objects are presented from unusual perspectives. We then conducted ablation studies to pin down the core computational mechanisms underlying 3D object recognition.

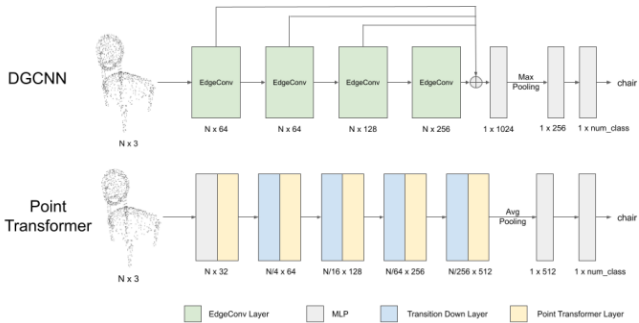

Figure 1. The architectures of DGCNN (top) and Point Transformer (bottom). MLP: Multi-layer perception, consisting of multiple fully connected layers. ⊕: concatenation.

## Modeling Methods

### Dataset

We used stimuli selected from the publicly available point cloud dataset, ModelNet40 dataset (Wu et al., 2015). The ModelNet40 dataset includes a set of 3D CAD models from 40 object categories. There are 12,311 3D CAD objects in total in the ModelNet40 dataset, where 9,843 objects are used for training, and 2,468 objects are used for testing. The point clouds consist of 1,024 points sampled uniformly from the surface of the 3D CAD models.

### Deep Learning Models

We evaluated two deep learning models for 3D object recognition from point cloud data: a convolution-based model (DGCNN; Wang et al., 2019) and a transformer-based model (Point Transformer; Zhao et al., 2021). Both models take 3D coordinates of points sampled from an object and generate a feature embedding for object classification.

DGCNN processes point clouds as graphs using EdgeConv layers, which extract local geometric features by comparing each point to its neighbors in feature space. Unlike traditional CNNs that operate on regular grids, DGCNN dynamically updates the neighborhood structure in each layer, enabling it to capture complex 3D shapes through local feature aggregation and global max pooling. Our implementation used 1024 points and 20 nearest neighbors, based on the publicly available pretrained model.

Point Transformer adopts a self-attention mechanism inspired by transformer architectures in NLP and vision. Each Point Transformer layer integrates information from neighboring points using attention weights based on spatial and feature similarity. Transition Down layers perform hierarchical downsampling, enabling the model to capture both local and global shape features. This design allows for adaptive focus on informative regions of the input, enhancing recognition performance.

While both models operate on point cloud inputs, DGCNN emphasizes local geometric structure, whereas Point Transformer combines hierarchical pooling and attention to capture context-dependent relationships. To align with human experiments, we trained both models on the ModelNet40 dataset and extracted logits corresponding to the ten object categories shown to participants, enabling a direct comparison of recognition performance. Both models were trained using the same data augmentation procedures, including random point dropout, random scaling, and random shifting of the input point clouds.

| **DGCNN** | **Point transformer** |
|---|---|
| Feature embeddings of a reference point are modulated by adding the weighted feature differences from the neighbor points | Feature embeddings of a reference point are computed using feature embeddings of neighbor points weighted by similarity |
| Neighbor points are defined in feature space, changing from layer to layer | Neighbor points are defined based on distance in 3D space |
| No spatial resolution change | Downsampling: spatial resolution change from fine to coarse |
| No position (3D coordinates) encoding | Position encoding added to each layer |

Table 1. Comparison and key differences between a convolution-based model (DGCNN) and transformer-based model (Point transformer).

## Experiments

### General Human Study Procedure

We use the same general procedure across all experiments unless otherwise specified.

Participants were instructed that they would view a rotating point cloud object during each trial. The stimulus was displayed for 3 seconds, after which ten buttons, each labeled with a different object name, appeared for selection. Their task was to select the object category that best matched the presented point cloud object. The ten object categories were airplane, bottle, bowl, chair, cup, lamp, person, piano, stool, and table.

Participants first completed a practice trial showing a rotating point cloud of a plant. They had to select the correct object category before proceeding to the experimental trials. If participants selected a wrong object category during practice, then the trial was repeated, and a hint message was displayed below the ten category buttons, directing participants to select the "Plant" button. This practice trial aimed to familiarize participants with the point-cloud display and ensure they understood the recognition task. The object category in the practice trial was not included in the subsequent experimental trials.

The experimental trials were similar to the practice trial, except that no feedback was provided. At the end of the experiment, demographic information was collected, and participants were presented with debriefing information about the study.

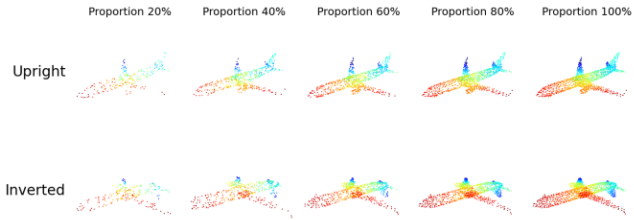

Figure 2. Downsampled and inverted point cloud stimuli used in Experiment. A random subset of points from the point cloud is retained in varying proportions. Colors represent depth, with red indicating proximity and blue indicating distance. In the actual experiments, the stimuli were presented as black points with rotation in depth, viewed from a horizontal viewpoint.

## Experiment 1: Point density and object orientation

In the first experiment, we investigated the recognition performance of both human participants and neural network models under two conditions: (1) varying point cloud sparsity and (2) presentation of point clouds in an inverted (upside-down) orientation. By combining these two conditions, we examined the robustness of human and model recognition across different levels of detail and atypical viewpoints.

**Participants** Two groups of participants were recruited through the UCLA Subject Pool. For the sparse point cloud condition, 56 participants were recruited (45 female, 11 male), with one participant excluded for reporting a lack of seriousness, resulting in a final sample of 55 participants (mean age = 19.8, SD = 1.4). For the inverted point cloud condition, 47 participants were recruited (40 female, 7 male), with a mean age of 20.4 years (SD = 1.5). The average completion time for both conditions was approximately 10 minutes, with a slight variance between groups.

**Stimuli** The stimuli were selected from the test set of the ModelNet40 dataset to ensure a fair comparison between human participants and deep neural network (DNN) models. We selected 7 objects from each of the ten categories, where each stimuli was transformed under two conditions 1) sparse point clouds: each point cloud were randomly downsampled to seven proportions (20%, 30%, 40%, 50%, 60%, 80%, and 100% of 1024 points). 2) Inverted point clouds: each stimulus from condition 1 were flipped upside down. Therefore, we have 7 objects x 7 downsampling ratios x 10 categories = 490 stimuli for the upright condition and the inverted condition. Each point cloud was presented as a GIF rotating 10 degrees per frame around the vertical axis. The GIF is displayed at 10 frames per second, completing a full 360-degree rotation in 3.6 seconds.

In the experiment, the participants are split into two groups, where one group viewed the upright point clouds and the other viewed the inverted point clouds. Each participant viewed one object exemplar only once, with a random permutation of proportions. In other words, each participant viewed all seven objects from one category, and each object was displayed in the point cloud format with a different downsampling proportion. This resulted in a total of 70 trials per participant. The trials were randomized for each participant. For the models, we used all 490 objects x 2 conditions = 980 objects for testing.

**Results** The results of the experiment are presented in Figure 3. Despite the sparse information provided in point cloud display, human participants consistently demonstrated high accuracy across all levels of point density. Their performance ranged from 86.2% to 95.3%, with only a slight decline as the point density decreased. For instance, when the point clouds were downsampled to 20% of the original points, the mean accuracy was 86.4% (CI = [83.5%, 89.2%]), and at 30%, the mean accuracy was 88.9% (CI = [86.3%, 91.5%]). This suggests that human participants are highly resilient to reduced point density, maintaining reliable recognition performance even with sparsely represented 3D objects.

The performance of the DGCNN model, however, was markedly affected by the reduction in point density. While the model's accuracy approached human performance at higher point densities (above 30%), it declined sharply at lower proportions. Specifically, the accuracy dropped to 64.3% at 30% point density and 48.6% at 20%. In contrast, the Point Transformer model exhibited greater robustness to sparsity. Its accuracy remained high across all levels of point density, ranging from 94.3% at 50% to 87.1% at 20%, showing similar performance robustness as humans.

In the inverted condition, human participants showed the inversion effect with lower accuracy than upright condition, but still significantly greater than the chance level.

Furthermore, humans consistently outperformed the machine learning models, highlighting their adaptability to changes in perspective. Note that the Point Transformer model performed better than the DGCNN model overall and also showed less reduction in the inverted condition, particularly at lower point densities.

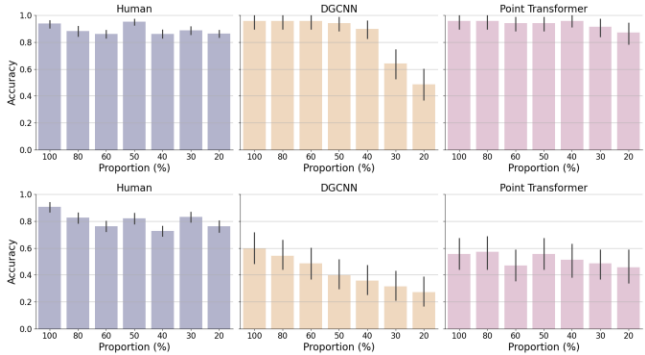

Figure 3. Accuracy of human participants and models as a function of point density for upright point clouds (top) and inverted point clouds (bottom). Error bars represent the 95% confidence interval.

### Experiment 2: Lego-Like Point Clouds

In Experiment 2, we aimed to disrupt the local geometric features of point clouds while preserving their global shape of 3D objects. This was achieved by converting point clouds into voxel grid displays, analogous to reducing the resolution of an image. The larger the voxel size, the lower the spatial resolution of the point cloud. The process involved generating a voxel grid from the point cloud, sampling points on the voxel surfaces, and normalizing the sampled points. The stimuli section below details the methodology and the corresponding implementation.

The idea of introducing Lego-like point clouds was inspired by the sawtooth images from Baker et al. (2018), where sawtooth effects were added to silhouette images to disrupt local contour features. In our 3D point cloud display, we similarly aimed to disrupt local curvature features by converting point clouds into Lego-like representations while keeping the global shape almost unchanged.

**Participants** A total of 60 participants were recruited through the university's Subject Pool. The sample comprised 49 females, 9 males, 1 non-binary individual, and 1 participant who preferred not to disclose their gender. The mean age of the participants was 20.6 years (SD = 3.9). The average completion time for the experiment was 6.3 minutes (SD = 2.3).

**Stimuli** We first converted each point cloud into a voxel grid representation using the Open3D library. A point cloud consists of discrete data points capturing an object's surface geometry, while voxels are small cubic units dividing 3D space into a regular grid, analogous to pixels in 2D images but extending into the third dimension. To convert a point cloud into a voxel grid, we superimposed a voxel grid over the point cloud, determine voxel occupancy by checking which voxels contain points, and then created a structured, blocky representation of the object.

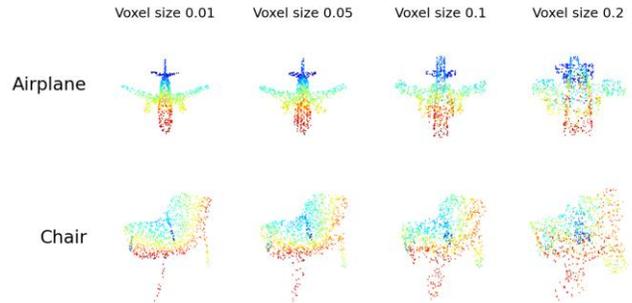

Figure 4. Lego-like Point Cloud Stimuli for Experiment 2. Points were uniformly sampled from voxel surfaces converted from point clouds, with varying voxel sizes determining resolution.

After obtaining the voxel grid representation of the point clouds, we sampled points uniformly on the surface of this grid. These sampled points were aggregated to form a new point cloud representing the voxel grid. Varying the voxel size allowed us to sample the point cloud at different resolutions, with larger voxel sizes resulting in lower resolution.

The resulting point cloud from the voxel sampling was then normalized to center it at the origin and scale it to fit within a unit sphere. This normalization does not affect the GIFs presented to human participants but is crucial for ensuring consistent input for machine learning models.

For the stimuli, we used the same 70 objects from 10 categories as in Experiment 1. Each object was sampled with four different voxel sizes: 0.01, 0.05, 0.1, and 0.2. Each participant viewed four random but non-overlapping objects from each voxel size within each category, constituting a total of 40 stimuli per participant.

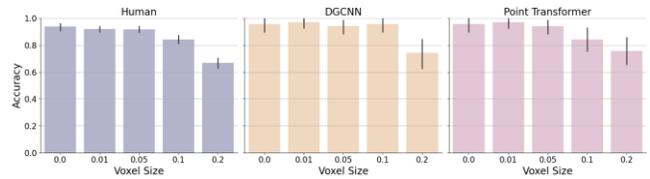

Figure 5. Accuracy of human participants and models on Lego-like point clouds with varying voxel sizes. Error bars represent the 95% confidence interval.

**Results** The results of Experiment 2 are depicted in Figure 5. Human participants demonstrated relatively stable accuracy across different voxel sizes. Human performance remained almost the same when voxel size increased from 0.0 (mean accuracy 93.8%, CI = [91.8%, 95.8%]) to 0.05 (mean accuracy 91.8%, CI = [89.5%, 94.1%]). Then accuracy gradually reduced as the voxel size increased to 0.1 (mean accuracy 84.3%, CI = [81.3%, 87.3%]) and 0.2 (mean accuracy 66.8%, CI = [62.9%, 70.7%]). This suggests that human participants can effectively recognize objects even when local geometric features are jittered, provided the global shape remains intact.

Both models achieved performance levels comparable to human observers at lower voxel sizes, suggesting their ability to generalize well with minimal distortion. However, as voxel size increased, performance began to decline across all groups. Notably, the Point Transformer exhibited a gradual decline in accuracy that closely mirrored the human trend, suggesting that it captures a similar sensitivity to degradation in spatial resolution. In contrast, DGCNN maintained stable performance up to voxel size 0.1 but showed a sharp drop at 0.2, indicating a threshold-like effect leading to brittle performance. Specifically, DGCNN's accuracy dropped from 95.71% at voxel size 0.1 to 74.29% at voxel size 0.2.

We further analyzed the correlation between model and human accuracy patterns across all experimental conditions, in the next section titled "Model and Human Performance Correlation" (also see Figure 8). This section provides a more detailed quantitative comparison of model alignment with human behavior across the two experiments.

## Investigating Mechanisms Underlying Global Shape Bias

To identify which computational mechanism in the Point Transformer model contribute to its global shape bias, we systematically evaluated three primary differences between the Point Transformer and the DGCNN: (1) the Attention mechanism, (2) Position Encoding, and (3) the Downsampling supporting hierarchical pooling and abstraction of 3D shapes. By selectively removing each component, we developed multiple variants of the Point Transformer and assessed their performance on the same downsampled point cloud stimuli used in Experiment 1.

Figure 6 illustrates the accuracy of different Point Transformer variants as a function of point density, stimuli used in Experiment 1. The original Point Transformer model, the "Original" variant in the plot, achieves the highest performance across all proportions. The "NoAttn" and "NoPE" variants, which removes the attention mechanism and the position encoding mechanism respectively, experiences a mild accuracy drop compared to the original. This suggests that while attention and position encoding are beneficial, the model maintains substantial effectiveness even in their absence.

In contrast, the variant lacking the downsampling mechanism ("NoDS") demonstrated performance comparable to the original model at higher point densities but exhibited significant accuracy declines at lower point densities. This result underscores the critical role of the downsampling component in generalizing across varying degrees of point sparsity. Downsampling facilitates hierarchical representation within the model, enabling a global integration of shape information and reducing dependence on local features.

Interestingly, this effect mirrors the visual pathways in the human brain, particularly the hierarchical processing in ventral pathway, indicated by increased receptive fields of neurons from the low-level to high-level visual areas. Similarly, downsampling in Point Transformer enforces a global shape bias, helping the model recognize objects even when the number of available points is significantly reduced. By progressively reducing the number of points covering the entire objects, downsampling allows the model to integrate local features into a coherent global representation, making it more robust to various data transformations.

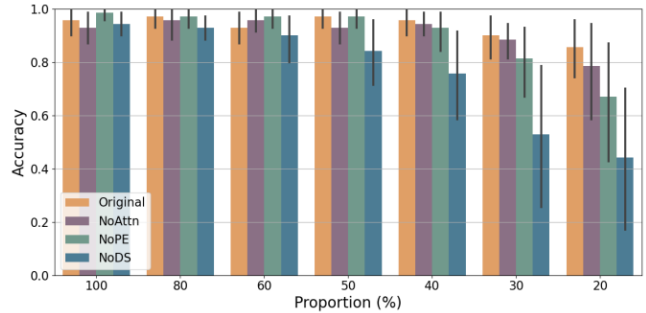

Figure 6. Accuracy of variants of Point Transformers on downsampled point clouds. Original = the original Point Transformer, NoAttn = No attention, NoPE = No position encoding, NoDs = No downsampling.

## Enhancing DGCNN with Downsampling

Given the critical role identified for the downsampling mechanism in fostering a global shape bias in the point transformer model, we next examined whether integrating this computational mechanism into the DGCNN architecture could enhance its performance. We introduced the Transition Down layer from the Point Transformer into each EdgeConv layer of the DGCNN, matching the structural depth of the original Point Transformer.

As demonstrated in Figure 7, incorporating the Transition Down layer significantly improved the DGCNN's robustness to point density. The original DGCNN exhibited significant performance degradation at point densities less than 40% . In contrast, the modified DGCNN with the Transition Down layer ("DGCNN + DS") increased the the accuracy for the low point density condition. For example, performance increased from 0.486 to 0.829 for the 20% point density, aligning closely with both human performance and the original Point Transformer.

The addition of the Transition Down layer effectively compels the model to construct more abstract, hierarchical representations of 3D objects, shifting its focus towards global shape characteristics rather than relying excessively on detailed local features. Crucially, this demonstrates that downsampling can effectively bridge the gap between seemingly distinct model architectures, transformer-based and convolution-based, highlighting a shared computational mechanism toward robust shape recognition. While it is commonly assumed that the attention mechanism in transformer-based architectures primarily enables better generalization, our results clearly demonstrate that it is actually the downsampling mechanism driving this effect. The success of downsampling across both transformer and convolution-based models challenges conventional assumptions about their inherent differences, suggesting that hierarchical abstraction strategies may universally benefit deep learning models.

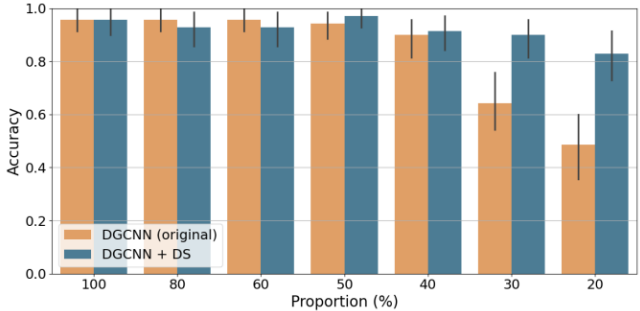

Figure 7. Accuracy of variants of DGCNN model on downsampled point clouds. DGCNN + DS = DGCNN model with additional Downsampling layer after each EdgeConv layer.

### Model and Human Performance Correlation

We compared the accuracy patterns of the DGCNN and Point Transformer, along with their respective variants, DGCNN + DS and Point Transformer noDS, to human accuracy patterns across experimental conditions in the two experiments. By pooling performance data across all tested conditions, we computed Pearson correlations between each model's accuracies and human responses.

As shown in Figure 8, both model variants that incorporate the downsampling mechanism (DGCNN + DS and Point Transformer) exhibited substantially higher correlations with human performance than their non-downsampling counterparts (DGCNN and Point Transformer noDS). Specifically, the correlation between human performance and DGCNN + DS was $r = 0.721$ ($p < 0.001$), and for the original Point Transformer it was $r = 0.711$ ($p < 0.001$), both significantly higher than the correlation with the original DGCNN ($r = 0.545$, $p = 0.016$). These differences were statistically supported by comparisons such as DGCNN vs. DGCNN + DS ($z = -15.570$, $p < 0.0001$) and DGCNN vs. Point Transformer ($z = -8.163$, $p < 0.0001$), indicating that the introduction of downsampling yields not only improved model robustness but also closer alignment to human accuracy profiles.

This finding provides converging evidence that downsampling plays a central role in eliciting human-like recognition of 3D shapes. The enhanced human-model correspondence across two structurally distinct architectures further underscores the broad applicability of this mechanism.

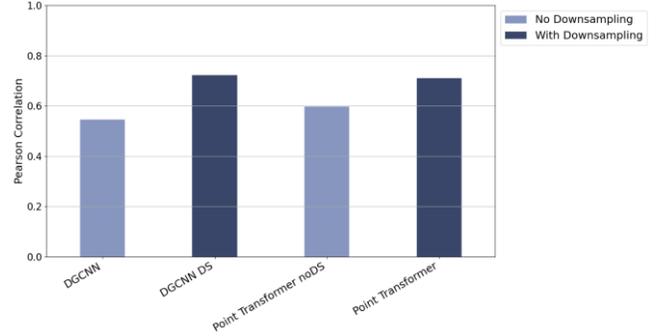

Figure 8. Correlation between model and human accuracy patterns across conditions. Models with downsampling show significantly stronger alignment with human.

### Conclusions and Discussions

Across two behavioral experiments and a series of modeling analyses, we demonstrated that humans exhibit strong robustness in recognizing 3D objects, even under challenging conditions such as sparse input, inversion, or disrupted local geometry. While current deep learning models—DGCNN and Point Transformer—achieved comparable accuracy to humans under standard conditions, their robustness under distortion varied substantially.

In particular, the Point Transformer model consistently mirrored human performance trends, showing gradual declines under increasing point sparsity and local geometry disruption. In contrast, the DGCNN model was more brittle, with sharp drops in performance in the untrained conditions. These findings indicate that models relying predominantly on local geometric features, such as DGCNN, lack the holistic processing strategies characteristic of human perception.

Our ablation studies further revealed that the downsampling mechanism in Point Transformer is the key component supporting its robustness and global shape bias in 3D object recognition. By introducing this mechanism into DGCNN, we significantly improved its performance and alignment with human responses. Importantly, our findings challenge the common assumption that attention mechanisms are the main contributors to strong generalization in transformer-based models; instead, hierarchical abstraction through downsampling plays the more critical role for 3D shape recognition.

Taken together, our results highlight the importance of global shape representations in achieving human-like 3D object recognition. Integrating cognitively inspired mechanisms, such as hierarchical downsampling, into deep learning models can improve their generalization and robustness, bringing them closer to the perceptual strategies employed by the human visual system.

### Acknowledgments
This work was supported by the NSF grant BCS-2142269.

# References

Baker, N., Lu, H., Erlikhman, G., & Kellman, P. J. (2018). Local features and global shape information in object classification by deep convolutional neural networks. *Vision Research*, 159, 60-72.

Biederman, I. (1987). Recognition-by-components: A theory of human image understanding. *Psychological Review*, 94(2), 115-147.

Dosovitskiy, A., Beyer, L., Kolesnikov, A., Weissenborn, D., Zhai, X., Unterthiner, T., Dehghani, M., Minderer, M., Heigold, G., Gelly, S., Uszkoreit, J., & Houlsby, N. (2020). An Image is Worth 16x16 Words: Transformers for Image Recognition at Scale. *Proceedings of the International Conference on Learning Representations (ICLR)*.

Geirhos, R., Rubisch, P., Michaelis, C., Bethge, M., Wichmann, F. A., & Brendel, W. (2018). ImageNet-trained CNNs are biased towards texture; increasing shape bias improves accuracy and robustness. *arXiv preprint* arXiv:1811.12231.

Goldstone, R. L. (1998). Perceptual learning. *Annual Review of Psychology*, 49(1), 585-612.

Guo, Y., Wang, H., Hu, Q., Liu, H., Liu, L., & Bennamoun, M. (2020). Deep learning for 3D point clouds: A survey. *IEEE Transactions on Pattern Analysis and Machine Intelligence*, 43(12), 4338-4364.

He, K., Zhang, X., Ren, S., & Sun, J. (2016). Deep residual learning for image recognition. *Proceedings of the IEEE Conference on Computer Vision and Pattern Recognition (CVPR)*.

Hoffman, D. D., & Singh, M. (1997). Salience of visual parts. *Cognition*, 63(1), 29-78.

Kellman, P. J. (1984). Perception of three-dimensional form by human infants. *Perception & Psychophysics*, 36, 353-358.

Krizhevsky, A., Sutskever, I., & Hinton, G. E. (2012). ImageNet classification with deep convolutional neural networks. *Advances in Neural Information Processing Systems*, 25, 1097-1105.

Liu, Z. (1998). Ideal observers of visual object recognition. Theoretical Aspects of Neural Computation, *A Multidisciplinary Perspective*, 145-154.

Marr, D. (1982). Vision: A computational investigation into the human representation and processing of visual information. *MIT Press*.

Murray, S. O., Olshausen, B. A., & Woods, D. L. (2003). Processing shape, motion and three-dimensional shape-from-motion in the human cortex. *Cerebral cortex*, 13(5), 508-516.

Palmer, S. E. (1999). Vision science: Photons to phenomenology. *MIT Press*.

Qi, C. R., Su, H., Mo, K., & Guibas, L. J. (2017a). PointNet: Deep learning on point sets for 3D classification and segmentation. *Proceedings of the IEEE Conference on Computer Vision and Pattern Recognition*, 652-660.

Qi, C. R., Yi, L., Su, H., & Guibas, L. J. (2017b). PointNet++: Deep hierarchical feature learning on point sets in a metric space. *Advances in Neural Information Processing Systems*, 30, 5099-5108.

Shepard, R. N., & Metzler, J. (1971). Mental rotation of three-dimensional objects. *Science*, 171(3972), 701-703.

Szegedy, C., Zaremba, W., Sutskever, I., Bruna, J., Erhan, D., Goodfellow, I. J., & Fergus, R. (2014). Intriguing properties of neural networks. *Proceedings of the International Conference on Learning Representations (ICLR)* (Poster).

Treue, S., Husain, M., & Andersen, R. A. (1991). Human perception of structure from motion. *Vision research*, 31(1), 59-75.

Treue, S., Andersen, R. A., Ando, H., & Hildreth, E. C. (1995). Structure-from-motion: Perceptual evidence for surface interpolation. *Vision research*, 35(1), 139-148.

Vaswani, A., Shazeer, N., Parmar, N., Uszkoreit, J., Jones, L., Gómez, A. N., Kaiser, Ł., & Polosukhin, I. (2017). Attention is all you need. *Advances in Neural Information Processing Systems (NeurIPS)*, 30, 5998–6008.

Wagemans, J., Elder, J. H., Kubovy, M., Palmer, S. E., Peterson, M. A., Singh, M., & von der Heydt, R. (2012). A century of Gestalt psychology in visual perception: I. Perceptual grouping and figure-ground organization. *Psychological Bulletin*, 138(6), 1172-1217.

Wallach, H., & O'connell, D. N. (1953). The kinetic depth effect. *Journal of experimental psychology*, 45(4), 205.

Wang, Y., Sun, Y., Liu, Z., Sarma, S. E., Bronstein, M. M., & Solomon, J. M. (2019). Dynamic Graph CNN for learning on point clouds. *ACM Transactions on Graphics* (TOG), 38(5), 1-12.

Wu, Z., Song, S., Khosla, A., Yu, F., Zhang, L., Tang, X., & Xiao, J. (2015). 3D ShapeNets: A deep representation for volumetric shapes. *Proceedings of the 28th IEEE Conference on Computer Vision and Pattern Recognition (CVPR)*.

Yee M, Jones SS and Smith LB (2012) Changes in visual object recognition precede the shape bias in early noun learning. *Front. Psychology,* 3, 533.

Zhao, H., Jiang, L., Jia, J., Torr, P. H., & Koltun, V. (2021). Point transformer. *Proceedings of the IEEE/CVF international conference on computer vision* (pp. 16259-16268).
7